\documentclass[11pt,a4paper]{article}
\usepackage[hyperref]{emnlp2020}
\usepackage{times}
\usepackage{latexsym}
\usepackage{tabularx}

\usepackage{algorithm}
\usepackage{algorithmicx}
\usepackage{diagbox}
\usepackage[noend]{algpseudocode}
\usepackage{multirow}
\usepackage{amsmath}
\usepackage{graphicx}
\usepackage{hhline}
\usepackage{amsmath}
\usepackage{caption}
\usepackage{epstopdf}
\usepackage{cleveref}
\usepackage{soul}
\usepackage{subcaption}
\usepackage[normalem]{ulem}
\usepackage{booktabs}
\usepackage{kotex}
\DeclareSymbolFont{matha}{OML}{txmi}{m}{it}
\DeclareMathSymbol{\varv}{\mathord}{matha}{118}

\setstcolor{red}
\newcommand{\RNum}[1]{\expandafter{\romannumeral #1\relax}}

\DeclareMathOperator*{\argmax}{arg\,max}

\newcommand{\abs}[1]{\lvert#1\rvert}
\usepackage[super]{nth}
\usepackage{color}
\usepackage{xcolor}

\definecolor{lightblue}{rgb}{.8,.95,1}
\definecolor{LightBlue}{rgb}{.8,.9,1}
\sethlcolor{lightblue}
\definecolor{green}{RGB}{131, 201, 55}
\definecolor{cadmiumorange}{rgb}{0.93, 0.53, 0.18}
\definecolor{cadmiumyellow}{RGB}{203, 180, 11}
\definecolor{purple}{RGB}{70, 10, 100}
\definecolor{cadmiumred}{rgb}{1.0, 0.0, 0.22}
\definecolor{blue}{rgb}{0.19, 0.55, 0.91}
\definecolor{tomato}{rgb}{1.0, 0.39, 0.28}
\definecolor{fuchsiapink}{RGB}{170, 100, 170}

\usepackage{microtype}

\aclfinalcopy 
\def\aclpaperid{334} 

\title{F$^{2}$-Softmax: Diversifying Neural Text Generation via \\ Frequency Factorized Softmax}

\author{Byung-Ju Choi\textsuperscript{1} \and  Jimin Hong\textsuperscript{1} \and David Keetae Park\textsuperscript{2} \and Sang Wan Lee\textsuperscript{1,3,4} \\
        \textsuperscript{1}Humelo, Republic of Korea \\
        \textsuperscript{2}Department of Biomedical Engineering, Columbia University, USA \\
        \textsuperscript{3}Department of Bio and Brain Engineering, KAIST, Republic of Korea \\
        \textsuperscript{4}Center for Neuroscience-inspired AI, KAIST, Republic of Korea \\
        \tt{\{bjej1123, jimin9401\}@gmail.com} \\
        \tt{dkp2129@columbia.edu, sangwan@kaist.ac.kr}} 
        
\date{}

\begin{document}
\maketitle
\begin{abstract}

Despite recent advances in neural text generation, encoding the rich diversity in human language remains elusive. We argue that the sub-optimal text generation is mainly attributable to the imbalanced token distribution, which particularly misdirects the learning model when trained with the maximum-likelihood objective. As a simple yet effective remedy, we propose two novel methods, \textit{F$^2$-Softmax} and \textit{MefMax}, for a balanced training even with the skewed frequency distribution. \textit{MefMax} assigns tokens uniquely to frequency classes, trying to group tokens with similar frequencies and equalize frequency mass between the classes. \textit{F$^2$-Softmax} then decomposes a probability distribution of the target token into a product of two conditional probabilities of (\RNum{1}) frequency class, and (\RNum{2}) token from the target frequency class. Models learn more uniform probability distributions because they are confined to subsets of vocabularies. 
Significant performance gains on seven relevant metrics suggest the supremacy of our approach in improving not only the diversity but also the quality of generated texts.


\end{abstract}

\section{Introduction}
Neural text generation is one of the extensively studied tasks of natural language processing (NLP), as it forms the basis for dialogue systems~\cite{chen2017survey}, machine translation~\cite{chaudhary2018machine}, and text summarization~\cite{kryscinski2019neural}. However, often monotonous or dull, texts generated from existing methods do not fully reflect the rich diversity and expression in human language~\cite{welleck2019neural}. In particular, models tend to overproduce words frequently appearing in the data, while hardly utilizing informative words  \cite{dinan2020second}.
Even pre-training techniques on large corpora fail to resolve the issue~\cite{holtzman2019curious}.


Possible causes for text degeneration have been illuminated, such as a defect specific to model architectures~\cite{vig2018deconstructing} or the discrepancy between training data and a true distribution~\cite{holtzman2018learning, jiang2019improving}. Recently, the emphasis has been placed on investigating the flaws in the maximum likelihood objective~\cite{holtzman2019curious}. Concretely, the likelihood training pays little attention to the top ranks in terms of the target token probabilities~\cite{welleck2019neural}, or maximizing likelihood itself does not adequately reflect human language processing~\cite{holtzman2019curious}. Therefore, with the maximum likelihood-based training, models learn to produce tokens frequently appearing in the data more often. 

We argue, however, that the primary reason behind the sub-optimal performance of the likelihood objective is essentially the imbalanced token distribution inherent in natural language. Natural language is extremely skewed in distribution, where the top hundred most frequently-used (top-100) words occupy nearly half of the total corpus~\cite{fagan2011introduction} following the Zipf's law~\cite{zipf1949human}. Training a classifier with the inherently imbalanced data on the maximum likelihood estimation (MLE) leads to biased classification boundaries in favor of majority classes~\cite{khan2019striking}. In other words, models play a difficult role in learning with the imbalanced label (i.e., token) distribution~\cite{he2008improving}.



\begin{figure*}[t]
\centering
\includegraphics[width=\textwidth]{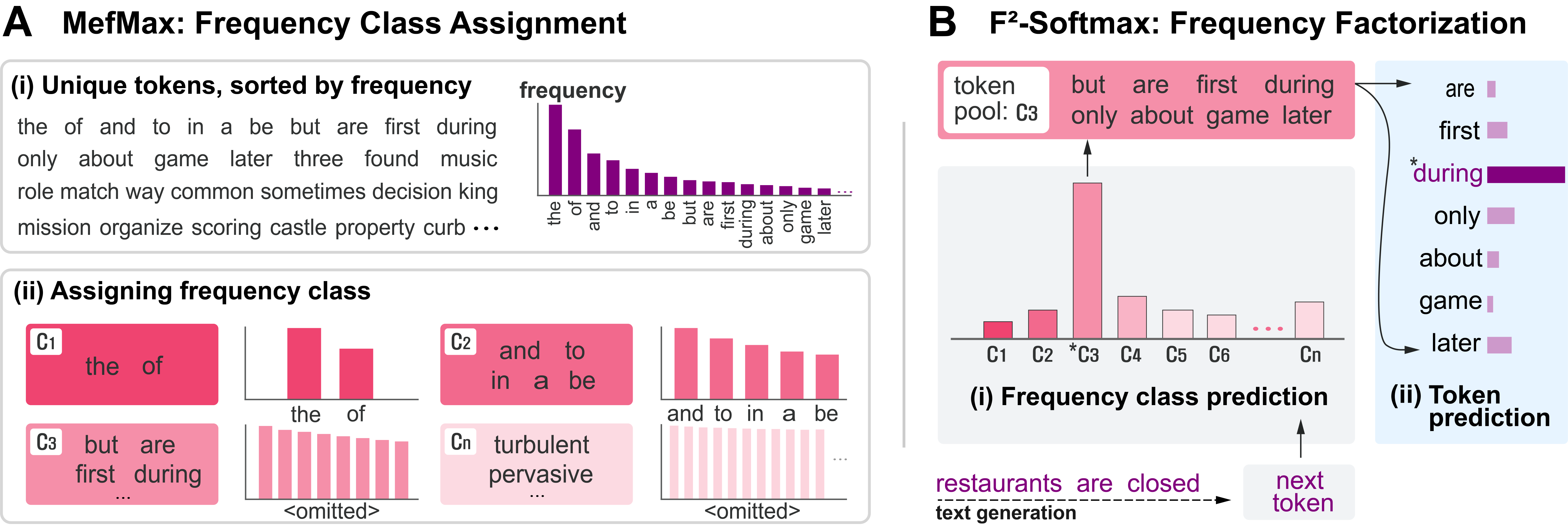}
\caption{Illustrations for working mechanisms of the two proposed modules, MefMax and F$^2$-Softmax. (A) Once unique tokens are sorted by frequency, MefMax (Section~\ref{sec:mme}) groups them to a set of frequency classes. The frequency distributions of the grouped tokens are more uniform than the distribution of the full vocabulary set. (B) With the frequency class assigned, token generation is decomposed into (i) predicting the frequency class, and (ii) generating the target token from the given frequency class.}
\label{fig:main}
\end{figure*}


We hypothesize that text generation can be enriched by balancing out the training data distribution. To this end, we introduce F$^2$-Softmax (Fig.~\ref{fig:main}(B), Section~\ref{sec:f2s}), which factorizes the probability distribution of the target token into a product of two conditional probabilities of (i) frequency class, and (ii) token from the target frequency class. It ensures training over balanced data, since the frequency classes are designed to have the distribution close to uniformity, and token distributions within a class are confined to  subsets of vocabularies grouped with similar frequencies. To this end, all unique tokens are assigned to a frequency class prior to the training, by our novel \textit{mean efficiency maximization} (MefMax; Fig.~\ref{fig:main}(A), Section~\ref{sec:mme}). MefMax evaluates and maximizes the class-labeling performance with the normalized entropy (i.e., efficiency), having the probability distributions to be learned as uniform as possible. 


We conduct extensive performance evaluations on seven relevant metrics that quantify the diversity and quality of generated texts. In terms of the diversity of generated texts, our approach significantly outperforms not only the MLE baseline~\cite{radford2019language} but also other diversity-promoting alternatives \cite{welleck2019neural,jiang2019improving}. We also achieve state-of-the-art results on most of the quality performances.
 
\section{Related Works}
\subsection{Diversity-promoting Text Generation}
In the field of neural text generation, prior studies either take a training-based approach or a decoding-based approach to promote the diversity in the generated texts.
\paragraph{Training-based Approach.} 
In dialogue generation, stimulating models to generate texts that share high mutual information with the contexts~\cite{li2016diversity} has shown to improve the diversity of output tokens by adding a maximum mutual information (MMI) constraint to the standard likelihood objective. Meanwhile, FACE~\cite{jiang2019improving} dynamically weights the cross-entropy losses based on target token frequencies, to prevent excessive weight-updates of some frequently used words. 
In another line of works for language modeling, text diversity has been promoted by a learning-to-cooperate framework in which multiple discriminators cooperate to reach a common goal~\cite{holtzman2018learning}. Also, the unlikelihood training strategy penalizes repetition with auxiliary loss terms \cite{welleck2019neural}. Such works are orthogonal to ours since F$^2$-Softmax focuses on decomposing the softmax function without employing an auxiliary loss or loss re-scaling.

\paragraph{Decoding-based Approach.} One of the widely used decoding tactics for promoting the diversity and richness of texts is stochastic decoding. Top-\textit{k} sampling stochastically samples the next token from the top-\textit{k} candidates in the predicted probability distribution~\cite{fan2018hierarchical}. Another pillar of stochastic decoding is nucleus sampling, which selects the next token from the top-\textit{p} portion of the probability mass~\cite{holtzman2019curious}. Other studies include beam blocking~\cite{paulus2017deep} in which the probabilities of tokens are set to zero if they were to create repeating n-grams, diverse beam search~\cite{vijayakumar2018diverse} which integrates dissimilarity terms into beam scores. Iterative beam search~\cite{kulikov2019importance} enhances diverse beam search with multiple iterations of beam search with different search spaces. These techniques are agnostic about model architecture or training methods. Our approach can be harmonically combined with the above techniques.

\subsection{Softmax Decomposition}
Decomposing the softmax function has long been studied in language modeling. Goodman (\citeyear{goodman2001classes}) decomposed the softmax function using a two-level hierarchy. This idea was generalized to deeper hierarchies in a later study~\cite{mnih2009scalable}. Approaches to construct softmax hierarchies have followed, such as utilizing word clusters obtained from \textit{k}-means algorithms \cite{le2011structured} or implementing Huffman coding with word frequencies \cite{mikolov2013efficient}. Furthermore, dynamic programming has been applied to obtain an optimal set of word classes with minimal computational costs for calculating the softmax function~\cite{zweig2013speed}. The same process has also been streamlined to fit into modern GPU environments~\cite{grave2017efficient}. These techniques bear a resemblance to ours for the use of softmax decomposition. However, our goal is fundamentally different: we aim to balance the data distribution in training, whereas previous approaches share the primary goal of reducing computational costs.

\subsection{Imbalanced Classification}
That we assign tokens to classes of balanced distribution shares a similar goal with overcoming imbalanced classification in computer vision domains. One of the widely adopted techniques for imbalanced classification is re-sampling, which includes removing examples from the majority classes (under-sampling) and adding samples for the minority classes (over-sampling) \cite{buda2018systematic}. Techniques for over-sampling include interpolating samples from neighboring samples~\cite{chawla2002smote} and adaptively synthesizing samples~\cite{he2008adasyn}. Cost-sensitive learning dynamically re-weights costs based on sample difficulties~\cite{dong2017class} or effective number of samples~\cite{cui2018large}. Other studies for the data imbalance problem consider metric learning~\cite{huang2016learning}, knowledge transfer~\cite{wang2017learning}, and Bayesian estimation~\cite{khan2019striking}.

\section{Methods}


\subsection{Maximum Likelihood}
The goal of language modeling is to learn a model $\hat{p}(\mathbf{x})$ which best describes a joint probability distribution $p(\mathbf{x})$, where $\mathbf{x} = [x_1,\ldots, x_T]$ is a sequence of tokens and $x_i \in \mathcal{V}$ is a token from a vocabulary set. In an auto-regressive manner, $p(\mathbf{x})$ can be factorized into a product of conditional probabilities of tokens; $p(\mathbf{x}) = \prod_tp(x_t|\mathbf{x_{<t}})$. A conventional approach for the training is to maximize log-likelihood of a sequence $\mathbf{x}$ as the following:
\begin{equation}
    \mathcal{L}_{\text{MLE}}(\hat{p}, \mathbf{x}) = \sum_{t=1}^{T}\log\hat{p}(x_t|\mathbf{x}_{<t}).
\end{equation}

\subsection{F$^2$-Softmax}
\label{sec:f2s}
We propose to factorize the posterior $\hat{p}(x_t|\mathbf{x_{<t}})$ into a product of two conditional probabilities:
\begin{equation}
\hat{p}_(x_t|\mathbf{x}_{<t}) = \hat{p}_{1}(c_t|\mathbf{x}_{<t})\times \hat{p}_{2}(x_t|c_t,\mathbf{x}_{<t}),
\label{fs}
\end{equation}
where $c_t \in \mathcal{C}$ denotes a frequency class label assigned to the token $x_t$ given the global frequency of the token in a corpus, belonging to a set of frequency classes $\mathcal{C}$. Following Eq. (\ref{fs}), the updated objective $\mathcal{L}_{F^2}(\hat{p})$ is then formulated as:
\begin{equation}
\begin{aligned}
\label{f^2fs}
&\mathcal{L}_{F^2}(\hat{p}, \mathbf{x})\\
&= \sum_{t=1}^{T}[\log\hat{p}_{1}(c_t|\mathbf{x}_{<t}) + \log\hat{p}_{2}(x_t|c_t,\mathbf{x}_{<t})].
\end{aligned}
\end{equation}
The objective is thus learning how to classify the target frequency of the token and selecting the exact token given the target frequency class.
The factorized probabilities $\hat{p}_{1}(c_t|\mathbf{x}_{<t})$ and $\hat{p}_{2}(x_t|c_t,\mathbf{x}_{<t})$ are defined empirically using softmax functions:
\begin{equation} \label{eq1}
\begin{split}
\hat{p}_{1}(c_t|\mathbf{x}_{<t}) & = \frac{\exp(\mathbf{h}_{t-1}\cdot\mathbf{u}^{c_t})}{\Sigma_{m \in \mathcal{C}}\exp(\mathbf{h}_{t-1}\cdot\mathbf{u}^{m})} \\
\hat{p}_{2}(x_t|c_t,\mathbf{x}_{<t}) & =\frac{\exp(\mathbf{h}_{t-1}\cdot\mathbf{o}^{x_t})}{\Sigma_{n \in \mathcal{V}_{c_t}}\exp(\mathbf{h}_{t-1}\cdot\mathbf{o}^{n})}, 
\end{split}
\end{equation}
where $\mathbf{h}_{t-1}$ is a hidden state of the context $\mathbf{x}_{<t}$; $\mathbf{o}^i$ and $\mathbf{u}^j$ can be viewed as output embedding vectors for $i \in \mathcal{V}_{c_t}$ and $j \in \mathcal{C}$, respectively, while $\mathcal{V}_{c_t}$ is a subset of vocabularies assigned to the class $c_t$. Note that $\hat{p}_{2}(x_t|c_t,\mathbf{x}_{<t})$ is computed from the narrowed pool of tokens $\mathcal{V}_{c_t}$ rather than the full vocabularies set $\mathcal{V}$. 
Since classes are differentiated based on the token frequency, tokens with the same class have similar frequencies. It ensures within-class frequency distribution of tokens is closer to uniform than that of the full vocabulary set.


\subsection{MefMax for Class Optimization}
\label{sec:mme}
The more uniform a label distribution is, the less likely decision boundaries are biased in favor of frequent classes. Therefore, we aim to maximize the degree of uniformity of frequency distributions for both (\RNum{1}) tokens within each class and (\RNum{2}) classes themselves (i.e., the sum of token frequencies within each class), to avoid the class imbalance problem~\cite{buda2018systematic} over the course of training. It is formalized as follows:

\begin{equation}
\label{argmax}
    \mathcal{C}^{'} = \argmax_\mathcal{C} [ \mathcal{U(\mathcal{C})} + \frac{1}{\abs{\mathcal{C}}}\sum_{i \in \mathcal{C}}\mathcal{U}(\mathcal{V}_{i})] ,
\end{equation}
where $\mathcal{U}$ is a function that measures the uniformity of the frequency distribution of a given set. While any tests of uniformity can be used as $\mathcal{U}$, we adopt \textit{Shannon's entropy}~\cite{shannon1948mathematical}. The entropy is a decent proxy for measuring the uniformity~\cite{dudewicz1981entropy}. 

\paragraph{Normalized Entropy.} Since the number of samples affects the entropy, entropy cannot be directly used. To marginalize the effect of the sample size, we use \textit{efficiency}, which is also known as the normalized entropy \cite{wijesekera1997shannon}, defined as:
\begin{equation}
    \mathcal{U(\mathbf{k})} = -\sum_{k_i \in \mathbf{k}}\frac{p(k_i)\log(p(k_i))}{\log (\abs{\mathbf{k}})}.
\end{equation}
It is equivalent to the ratio of the entropy to the maximum entropy, if the data were perfectly uniform. By applying the efficiency to Eq. (\ref{argmax}), our objective is to find a set of classes and their vocabularies such that their \textit{mean efficiency} is maximized.

\paragraph{Greedy Approach.}
The remaining issue is the computational overhead since the cost for exploring all possible class boundaries grows exponentially with the vocabulary size, not to mention the challenge of finding the optimal number of classes. To improve computational efficiency, we employ a straightforward greedy mechanism. It is based on the assumption that the mean efficiency is maximized when each class has approximately the same total frequency size. This assumption allows us to reduce our objective to optimizing the number of classes. Given a sorted vocabulary set $\mathcal{V}'$ and a candidate number of classes $K$, we divide classes so that each class has the same $1/K$ of total frequency. The optimal number of classes is the one that maximizes the mean efficiency. Algorithm~\ref{algo} shows the complete pseudo-code. Computation time is linear to the vocabulary size.

\begin{algorithm}[t]
\small
\caption{Pseudo-code for MefMax}\label{algo}

 \hspace*{\algorithmicindent} \textbf{Inputs :} Array $\mathcal{V}'$ of length $n$ sorted by the decreasing order of token frequency \\
 \hspace*{\algorithmicindent} \textbf{Outputs :} Number of classes, class boundary tokens 
\begin{algorithmic}[1]
\State $\mathcal{V}'$ $ \gets $ $\mathcal{V}'$ / sum($\mathcal{V}'$) \Comment{get relative frequencies}
\State maxMeanEfficiency $ \gets $ 0 
\State maxClassNum $ \gets 1 / \mathcal{V}'[0] $
\For{$K$ in [1, 2, ..., maxClassNum]}
        \State $B$ $\gets $ empty list \Comment{lists for candidate boundaries}
        \State tar $\gets 1/K $ \Comment{target frequency}
        \State cum, idx $\gets 0, 0 $ \Comment{cumulative frequency \& index}
        
        \While{tar $\leq$ 1} \Comment{compute candidate boundaries}
            \State cum $\gets$ cum + $\mathcal{V}'$[idx++]
        \If{cum $\geq$ tar}
            \State tar $\gets$ tar + $1/K$
            \State $B$.append(idx)
        
        \EndIf
        \EndWhile
    \State meanEfficency $\gets $ mean efficiency based on $B$
    \If{maxMeanEfficiency $<$ meanEfficency}
        \State maxMeanEfficiency $\gets$ meanEfficency
        \State $Out \gets B$
    \EndIf

\EndFor
\State \Return {len($Out$), $Out$}
\end{algorithmic}
\end{algorithm}

\subsection{Decoupled Decoding}
\label{sec:DD}
For the decoding stage, we decouple $\hat{p}_{1}$ from $\hat{p}$ in Eq.~(\ref{fs}) by first selecting a single frequency class from $\hat{p}_{1}$ and then generating the next token based on the selected class. For the target class $c_t=i$ sampled from the distribution $\hat{p}_1(c_t|\mathbf{x}_{<t})$, the probability for the next token is defined as:
\begin{equation}
  \hat{p}^{'}(x_t|\mathbf{x}_{<t}) =
    \begin{cases}
      \hat{p}_{2}(x_t|c_t=i, \mathbf{x}_{<t})  & \text{if } x_t \in \mathcal{V}_i\\
      0 & \text{otherwise}.
    \end{cases}       
    \label{dd}
\end{equation}
The target class can be sampled in both deterministic or stochastic manners, depending on decoding strategies. We found that the advantages of training balanced data distributions can be fully leveraged by sequentially performing tasks of frequency class prediction and token generation from the selected class.

\section{Experiments}

\subsection{Training and Evaluation Details}
\label{sec:experimental_details}
In this section, experimental details are illustrated. Exact hyperparameter settings and data statistics are described in Appendix.
\subsubsection{Datasets}
Two datasets that differ in language and text types are selected for the implementations. \\
\textbf{Wikitext-103} \footnote{\url{https://s3.amazonaws.com/research.metamind.io/wikitext/wikitext-103-v1.zip}} is a collection of English articles extracted from Wikipedia. Containing more than 100 million words, it is widely regarded as a benchmark dataset for language modeling. \\
\textbf{Melo-Lyrics} is a Korean lyrics dataset we crawled from multiple music streaming websites, including Soribada\footnote{\url{https://www.soribada.com}}, Genius\footnote{\url{https://genius.com}}, etc. Tokens in lyrics show a distribution largely different from general articles; for instance, repeated phrases are abundant in lyrics. Therefore it provides an additional unique angle for model evaluations and comparisons. It includes approximately 478 thousand songs with 51 million words in total.

\subsubsection{Model Architecture}
We use the Transformer \cite{vaswani2017attention}, an architecture well-suited for neural text generation \cite{lewis2019bart,welleck2019neural}. Specifically, we apply the Transformer decoder used in the GPT-2 model \cite{radford2019language}. Input texts are tokenized with the byte pair encoding \cite{sennrich2016neural}.

\subsubsection{Baseline Models}
For the baseline, we consider maximum likelihood estimation (MLE), a standard approach for text generation. Also compared are alternative models for promoting text diversities, including recently proposed FACE\footnote{\url{https://github.com/ShaojieJiang/FACE}} \cite{jiang2019improving} and unlikelihood training\footnote{\url{https://github.com/facebookresearch/unlikelihood_training}} (UL) \cite{welleck2019neural}. FACE improves text diversity by dynamically scaling losses, while the latter employs auxiliary losses.

\subsubsection{Training}
Training is carried out on a single GPU environment with 24GB of memory. We set all hyperparameters equal for all approaches by tuning them based on the validation losses of the MLE baseline for fair comparisons. We additionally optimize approach-specific hyperparameters of diversity-promoting baselines.

\subsubsection{Generation}
We generate texts for the evaluation by completing sequences from prefixes. Specifically, we batchify a test set, select the first 50 tokens from each batch as prefixes, and guide models to generate a continuation of 100 tokens from the prefixes. The experiments include both deterministic and stochastic decoding. We apply greedy search for deterministic decoding, and use top-\textit{k} sampling for stochastic decoding.


\begin{table*}[t]
\small
\centering
\begin{subtable}{\textwidth}
\centering
\begin{tabular}{l|l|l|lll|lll|lll|l|l}
\toprule[0.3ex]
                             \multirow{2}{0cm}{\diagbox[innerwidth=1.8cm]{Models}{Metrics}}        & \textbf{PPL} &  \textbf{KLD} & \multicolumn{3}{c|}{\textbf{MS-Jaccard}} & \multicolumn{3}{c|}{\textbf{Self-BLEU}} & \multicolumn{3}{c|}{\textbf{Distinct}} & \textbf{Rep} & \textbf{Uniq} \\

                             & & & \multicolumn{1}{c}{$n$=1}  & \multicolumn{1}{c}{$n$=2} &\multicolumn{1}{c|}{$n$=3}  & \multicolumn{1}{c}{$n$=1}  & \multicolumn{1}{c}{$n$=2} &\multicolumn{1}{c|}{$n$=3} &\multicolumn{1}{c}{$n$=1}  & \multicolumn{1}{c}{$n$=2} &\multicolumn{1}{c|}{$n$=3}&& \\
                             \toprule

\toprule
\textbf{MLE}      & \bf{24.7}  & 1.51 & 52.1& 35.6 & 24.3 & 93.4  &  83.2 &  69.7  & 45.1  & 71.9  & 83.0 & 0.67 & 8.48k \\
\textbf{FACE}     & 29.7  & 1.26 &  53.3& 33.2 & 21.7  & 92.3 & 76.5  & 57.6   & 53.4  & 77.1 & 85.5 & 2.1 & 10.3k         \\
\textbf{UL-token} & 25.8  & 1.28 &   54.3& 36.7 & 24.8 &  93.7 &     82.4   &  68.1 & 50.0   &  77.3 & 87.1 & 0.39 & 10.2k       \\
\textbf{UL-token+seq}  & 27.5 & 1.33  & 50.6 & 35.4 &  23.5  & \bf{95.3}    &  83.4  &  66.9   & 57.6    &  86.6 & 94.2  & \bf{0.11}  &  10.6k  \\
\hline
\rule{0pt}{2.4ex} \textbf{F$^{2}$-Softmax}   & 25.6 & \bf{0.62} &   \bf{67.4}&  \bf{42.4}  & \bf{26.4} &  93.3   &  \bf{71.9}  & \bf{48.1} &  \bf{65.7}   &  \bf{89.7}  & \bf{94.4}      & 0.33 & \bf{15.7k}    \\

\toprule
Human                         & - & -  & - &    - &   -   &   95.2 &  74.1   & 50.9  & 69.1& 92.1  & 95.8       & 0.11 & 15.2k  \\

\end{tabular}

\subcaption[]{Wikitext-103}
\vspace{0.3cm}
\end{subtable}
\begin{subtable}{\textwidth}
\centering
\begin{tabular}{l|l|l|lll|lll|lll|l|l}
\toprule[0.3ex]
                             \multirow{2}{0cm}{\diagbox[innerwidth=1.8cm]{Models}{Metrics}}        & \textbf{PPL} &  \textbf{KLD} & \multicolumn{3}{c|}{\textbf{MS-Jaccard}} & \multicolumn{3}{c|}{\textbf{Self-BLEU}} & \multicolumn{3}{c|}{\textbf{Distinct}} & \textbf{Rep} & \textbf{Uniq} \\
                             
                             & & & \multicolumn{1}{c}{$n$=1}  & \multicolumn{1}{c}{$n$=2} &\multicolumn{1}{c|}{$n$=3}  & \multicolumn{1}{c}{$n$=1}  & \multicolumn{1}{c}{$n$=2} &\multicolumn{1}{c|}{$n$=3} &\multicolumn{1}{c}{$n$=1}  & \multicolumn{1}{c}{$n$=2} &\multicolumn{1}{c|}{$n$=3}&& \\
                                     \toprule
                                     
\textbf{MLE}      & \bf{13.1}  & 0.34 & 67.6&47.3 & 32.4  & \bf{97.5}  &  83.1 &  62.2 &  56.3 & 75.5  & 84.0 &  1.1 & 22.4k \\
\textbf{FACE}     & 13.9  & 0.39 & 60.3 & 41.5 & 28.3  & \bf{97.5}  & 82.0    & 60.2  & 58.6  & \bf{78.1}  & 86.5  &  0.9      & 22.7k   \\
\textbf{UL-token} & 13.8  & 0.39 &62.4 &  43.6 & 29.6 & 98.2 &84.0  & 64.2 &  61.0   &  78.3 & 84.7  & 0.3 & 22.8k       \\
\textbf{UL-token+seq}  & 16.6 & 0.45   &58.3& 39.4 & 25.7  & 98.7 & 79.1   & 57.1   & 67.2   &  90.7  & 95.0  &   0.07 & 22.3k      \\
\hline
\rule{0pt}{2.4ex} \textbf{F$^{2}$-Softmax}   & 13.2 & \bf{0.13} &    \bf{78.8}& \bf{52.4} & \bf{34.4}  &\bf{97.5}   &  \bf{76.1} &   \bf{50.3} &  \bf{64.0}   &  78.7 &  \bf{83.9} & \bf{2.5} & \bf{25.2k}    \\
\toprule
Human                         & - & -  & - &    - &   -   &    97.5& 76.6  & 53.2 &  62.5 &  77.5   & 82.5      & 2.1 & 27.9k  \\

\end{tabular}
\subcaption[]{Melo-Lyrics}
\end{subtable}
\caption{Quantitative comparisons of F$^{2}$-Softmax with baseline models. Top-\textit{k} sampling strategy with the \textit{k} size of 3 and 20 are used for Wikitext-103 and Melo-Lyrics, respectively, across all models. UL-token imposes a token-level penalty, and UL-token+seq considers both token- and sequence-level penalties. PPL, KLD, and Rep are abbreviated notations for perplexity, KL-Divergence, and Repetition, respectively. Numbers $n\in (1,2,3)$ in column headers under MS-Jaccard, Self-BLEU, and Distinct refer to $n$-grams. Boldface scores denote the performances closest to humans. F$^{2}$-Softmax outperforms compared baselines in most of the cases. Results on various \textit{k} sizes are reported in Appendix.}

\label{main_res}
\end{table*}


\subsubsection{Evaluation Metrics}
\label{sec:evaluation metrics}
From seven total quantitative metrics we adopt to evaluate our model, Perplexity~\cite{bengio2003neural}, KL-Divergence~\cite{kullback1997information}, and MS-Jaccard~\cite{alihosseini2019jointly} are closely related to the \textit{likelihood} of generated texts. The other four metrics, namely Self-BLEU~\cite{zhu2018texygen}, Distinct-n~\cite{li2016diversity}, Repetition~\cite{holtzman2019curious}, and Uniq~\cite{welleck2019neural} measure the text \textit{diversity}. 
\\ \textbf{Perplexity}  quantifies the prediction difficulty over the next token. It is regarded as a general performance metric for text generation. 
\\ \textbf{KL-Divergence}  measures the difference between two probability distributions. We use unigram distributions of the generated texts and the test data. 
\\ \textbf{MS-Jaccard}  computes the similarity between the model's output and the ground truths by matching \textit{n}-grams.
\\ \textbf{Self-BLEU}  evaluates the inter-text diversity by computing BLEU \cite{papineni2002bleu} score for each generated text by regarding other outputs as reference.
\\ \textbf{Distinct-\textit{n}}  quantifies the intra-text diversity based on distinct \textit{n}-grams in each text. 
\\ \textbf{Repetition}  examines whether texts are stuck in repetitive loops. 
\\ \textbf{Uniq}  quantifies the richness of models using the number of unique generated tokens.

\subsection{Quantitative Comparisons}
\label{sec:quantitative_comparisons}
In this section, we report the scores computed from fully-trained models on the two benchmarks, Wikitext-103 and Melo-Lyrics, compared against baselines. Table~\ref{main_res} shows the results of stochastic decoding, while the results of deterministic decoding are reported in Table \ref{app:greedy_table}.

\subsubsection{Stochastic Decoding}
\paragraph{Wikitext-103.}
The desired qualities we aim for a text generation model is to generate human-like texts with a wide spectrum of token choices. Coupled with top-\textit{k} sampling, our F$^{2}$-Softmax achieves both goals by outperforming baselines with nearly all metrics compared, and closely approaching the human gold standard. 
As shown in Table~\ref{main_res}(a), our model is particularly effective in capturing the token diversity in the corpus. Notably, F$^{2}$-Softmax significantly improves both Self-BLEU and Distinct performances, having relative gaps to the human gold standard of 3.4\% and 3\%, respectively. The performance gaps of the second-best scores are 6.5\% (FACE) and 8.1\% (UL-token+seq), respectively. A surprising result is that F$^{2}$-Softmax improves Rep performance by 50\% over MLE, without an explicit penalty on repeating tokens. Another seminal contribution is the 30\% relative increase in unique tokens used for the generation, from the previously state-of-the-art level of 10.6k to 15.7k, as shown by the Uniq metric. This level closely reflects the human use of 15.2k tokens.

In PPL, which reflects the likelihood of the generated texts, the diversity-promoting baselines tend to perform worse than MLE, presumably due to the trade-offs between the diversity and the likelihood of texts. In contrast, F$^{2}$-Softmax maintains the smallest performance drop on PPL. F$^{2}$-Softmax also improves KLD and MS-Jaccard by 59\% and 19\% over MLE, respectively, which are large margins compared to the other comparatives.

\paragraph{Melo-Lyrics.}
Significant performance gains of F$^{2}$-Softmax are also observed in lyrics generation in Table~\ref{main_res}(b). The diversity-promoting baselines display severer degradation in PPL, KLD, and MS-Jaccard compared to the Wikitext-103 dataset. Especially, their repetition levels are significantly different from that of the ground truth data. We attribute this observation to the distinctive characteristics of lyrics, in which the same phrases are rhythmically repeated throughout the songs in the form of chorus or hook. Thus, for lyrics dataset, forcing models to discourage reusing previously used tokens may adversely affect the likelihood of the generated texts. Since F$^{2}$-Softmax helps models to diversify the output without an explicit regularization, models learn to generate well-thought-out tokens from the diversified token pool of 25.2k (Uniq), with state-of-the-art performances in KLD, MS-Jaccard, Self-BLEU, Distinct, and Rep.
\begin{table*}[ht]
\small
\centering
\label{app:stoc}
\begin{subtable}{\textwidth}
\centering
\begin{tabular}{l|l|l|lll|lll|lll|l|l}
\toprule[0.3ex]
                             \multirow{2}{0cm}{\diagbox[innerwidth=1.8cm]{Models}{Metrics}}        & \textbf{PPL} &  \textbf{KLD} & \multicolumn{3}{c|}{\textbf{MS-Jaccard}} & \multicolumn{3}{c|}{\textbf{Self-BLEU}} & \multicolumn{3}{c|}{\textbf{Distinct}} & \textbf{Rep} & \textbf{Uniq} \\

                             & & & \multicolumn{1}{c}{$n$=1}  & \multicolumn{1}{c}{$n$=2} &\multicolumn{1}{c|}{$n$=3}  & \multicolumn{1}{c}{$n$=1}  & \multicolumn{1}{c}{$n$=2} &\multicolumn{1}{c|}{$n$=3} &\multicolumn{1}{c}{$n$=1}  & \multicolumn{1}{c}{$n$=2} &\multicolumn{1}{c|}{$n$=3}&& \\
                             \toprule
                                     \toprule
\textbf{MLE}    & \bf{24.7} & 2.17 &   45.6& 29.9 &  20.5      &     92.8& 83.6   & 73.2 &  31.4 & 48.8  & 57.8  & 18.3 & 6.23k  \\
\textbf{FACE}  & 29.7 & 1.67 &  47.9 & 29.9 &  19.8  &   89.6& \bf{73.6} & \bf{57.3}  &  38.4& 54.3 & 62.0   & 21.0 & 8.03k   \\
\textbf{UL-token}    & 25.8 &  1.88 &    47.2 &30.7 & 20.9  &    92.9& 83.7   & 73.3 & 37.1&   56.3& 65.6 & 12.8 & 7.66k       \\
\textbf{UL-token+seq} & 27.5 & 2.06   &  41.5& 26.9 & 18.4 & \bf{95.6} & 86.6 & 74.2 & \bf{49.9}& \bf{78.1} & \bf{89.8} & \bf{0.3} & 8.33k      \\
\hline
\textbf{F$^{2}$-Softmax} & 25.6 &   \bf{1.63} &   \bf{49.0}& \bf{31.2} & \bf{21.0} & 90.2 & 78.7  & 66.3   & 36.3 &54.6 & 63.8 & 12.5 & \bf{9.08k}    \\

\toprule
Human                         & - & -  & - &    - &   -   &   95.2 &  74.1   & 50.9  & 69.1& 92.1  & 95.8       & 0.11 & 15.2k  \\

\end{tabular}
\subcaption[]{Wikitext-103}
\vspace{0.3cm}
\end{subtable}
\begin{subtable}{\textwidth}
\centering
\begin{tabular}{l|l|l|lll|lll|lll|l|l}
\toprule[0.3ex]
                             \multirow{2}{0cm}{\diagbox[innerwidth=1.8cm]{Models}{Metrics}}        & \textbf{PPL} &  \textbf{KLD} & \multicolumn{3}{c|}{\textbf{MS-Jaccard}} & \multicolumn{3}{c|}{\textbf{Self-BLEU}} & \multicolumn{3}{c|}{\textbf{Distinct}} & \textbf{Rep} & \textbf{Uniq} \\

                             & & & \multicolumn{1}{c}{$n$=1}  & \multicolumn{1}{c}{$n$=2} &\multicolumn{1}{c|}{$n$=3}  & \multicolumn{1}{c}{$n$=1}  & \multicolumn{1}{c}{$n$=2} &\multicolumn{1}{c|}{$n$=3} &\multicolumn{1}{c}{$n$=1}  & \multicolumn{1}{c}{$n$=2} &\multicolumn{1}{c|}{$n$=3}&& \\
                             \toprule
\textbf{MLE}    & \bf{13.1} & 0.46 &  64.5 & 44.2 & 31.6 & 91.4& 71.8 & 51.5 & 22.0& 26.7 & 28.5   & 67.7 & 20.4k  \\
\textbf{FACE}  & 13.9 & 0.51 & 57.7& 39.6  & 22.8   & 92.3 & 72.5 & 55.6  &25.3 & 32.0 & 36.1 & 39.1 & \bf{21.7k}       \\
\textbf{UL-token}    & 13.8 &  0.51 &    62.8 & 42.7 & 30.7  &  92.3 & \bf{73.0} & \bf{53.5} &  21.8 & 26.4 & 28.5 & 66.6 & 20k       \\
\textbf{UL-token+seq} & 16.6 & 0.74   &50.6 &  34.1  & 23.9 & \bf{95.7}&  78.7   & 56.3      &  \bf{29.6} & \bf{52.2} & \bf{57.1} & \bf{25.1} & 19.9k      \\
\hline
\textbf{F$^{2}$-Softmax}     & 13.2 &   \bf{0.38} & \bf{67.4} & \bf{45.1} & \bf{31.7}  &  90.4 & 66.6  & 43.1  & 21.5& 26.0  & 28.0& 66.9 & 21.3k    \\

\toprule
Human                         & - & -  & - &    - &   -   &    97.5& 76.6  & 53.2 &  62.5 &  77.5   & 82.5      & 2.1 & 27.9k  \\

\end{tabular}
\subcaption[]{Melo-Lyrics}
\end{subtable}
\caption{Evaluation results on the greedy sampling. The abbreviations are the same as Table \ref{main_res}.}
\label{app:greedy_table}
\end{table*}
\subsubsection{Deterministic Decoding}
In deterministic decoding, there is no clear method that outperforms the others in all of the metrics. For example, UL-token+seq exhibits the best performance in Distinct and Rep, while presenting the worst score in MS-Jaccard. Similarly, FACE improves Self-BLEU in exchange for performance loss on PPL and MS-Jaccard. Since we have seven metrics to compare, we conduct pair-wise evaluations between the compared methods, in which a method outperforms the other when a majority of metrics record higher. Our approach beats compared methods seven out of eight times (Table \ref{app:comp}). This result supports the supremacy of our approach regardless of the choice of decoding strategies.

However, deterministic decoding does not see the same amount of benefits obtained from stochastic decoding. We empirically find from our analyses that argmax operation in deterministic settings may harm the diversity when target class probabilities are nearly evenly distributed. We plan to delve deeper into our approach to improve our approach further.


\begin{figure}[t]
\includegraphics[width=.95\linewidth]{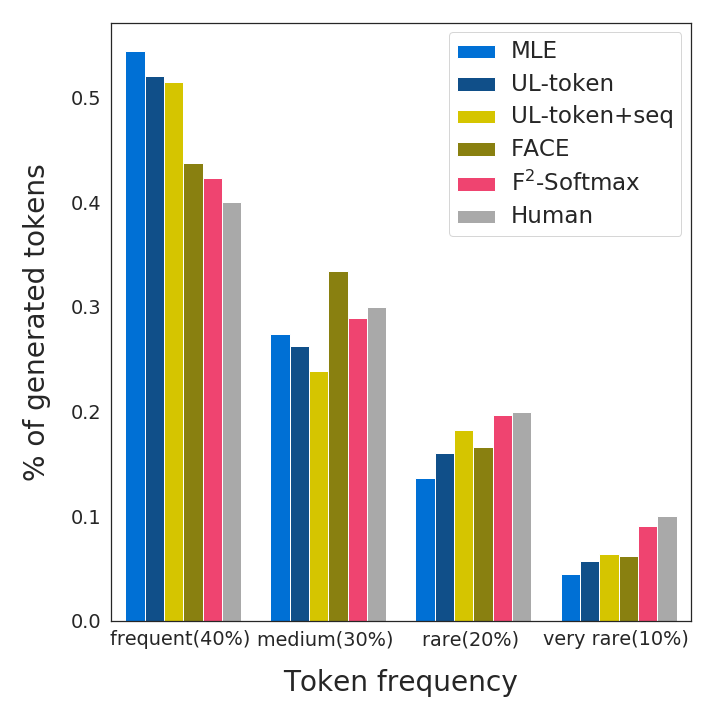}
\caption{Frequency distribution comparisons on the Wikitext-103 test set. Tokens in each group are defined based on the frequency mass of the training set. Tokens occupying the top 40\% of the frequency mass are assigned to \textit{frequent}, while those corresponding to the bottom 10\% are classified to \textit{very rare}.}
\label{fig:token distribution}
\end{figure}




\begin{figure*}[t]
\includegraphics[width=0.95\textwidth]{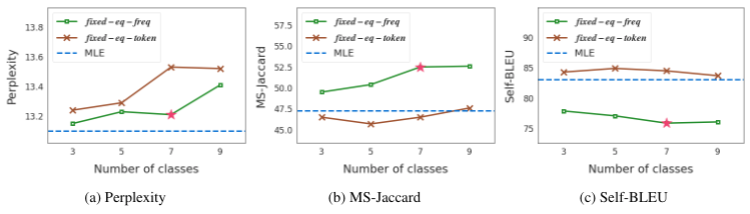}
\caption{Bigram performance on the Melo-lyrics test set with different numbers of classes. Green and brown lines indicate models with tokens are distributed based on frequency mass and token size, respectively. Blue dotted lines and star marks represent the MLE baseline and the choice of MefMax, respectively.}
\label{abl-MefMax}
\end{figure*}


\subsection{Learning Balanced Distribution}
\label{sec:Learning_Balanced_Distribution}

 The characteristic markers of monotonous texts are an overproduction of frequently used tokens and under-representation of rare tokens. To compare how models differentially generate tokens from frequent and rare tokens, we count the number of generated tokens corresponding to four defined categories of \textit{frequent}, \textit{medium}, \textit{rare} and \textit{very rare}. Tokens in each category are predefined from the Wikitext-103 training set. Fig.~\ref{fig:token distribution} plots the distribution results. MLE produces \textit{frequent} tokens 34\% more than human, while under-producing \textit{rare} and \textit{very rare} tokens by 40\%. The unlikelihood training baselines (UL-token, UL-token+seq) improve the diversity against MLE, but their results are relatively far from the real distribution. FACE manages to regulate a disproportionate use of \textit{frequent} tokens, but it fails to generate adequate amount of \textit{very rare} tokens. Generation results of our F$^{2}$-Softmax are closest to the gold standard, with the differences in \textit{frequent} and \textit{rare} tokens falling within 6\%.




\subsection{Ablation Studies}
\label{sec:Ablation}

In this section, we justify the pivotal roles of MefMax (Section~\ref{sec:mme}) and the decoupled decoding strategy (Section~\ref{sec:DD}). In order to assess contributions toward the final performances, we conduct a series of ablation tests. Stochastic decoding is used for the ablation studies.
\subsubsection{Ablation on MefMax} MefMax finds a desirable number of classes, intending to balance the frequency distribution of tokens between classes. Does MefMax help achieve better generation results than possible variants of class assignment? We answer this question by comparing the final performances against two simpler variants of MefMax. We name the first variant as \textit{fixed-eq-token}  in which tokens are distributed in \textit{equal} numbers to a \textit{fixed} number of classes. The second variant, \textit{fixed-eq-freq}, also assumes a fixed number of classes, but tokens are assigned to minimize the difference in the frequency distribution between classes. 

Fig.~\ref{abl-MefMax} presents the results. Clearly, \textit{fixed-eq-freq} outperforms \textit{fixed-eq-token}. It indicates that a decomposition of the softmax function without consideration of the data distribution (i.e., frequency distribution) aggravates both the likelihood and token diversity performances, regardless of the number of classes. For \textit{fixed-eq-token}, we find that models tend to overclassify the target class to the first class, which contains most of the total frequency, having most tokens generated from a fraction of the total vocabulary. This finding also justifies the hypothesis that balanced data distribution is an important factor in text generation.

Assigning classes based on the frequency (i.e., \textit{fixed-eq-freq}) continues to improve MS-Jaccard and Self-BLEU until the number of classes reaches the class choice of MefMax. With a larger number of classes than the choice, performances either plateau or decrease, demonstrating that MefMax is capable of selecting the optimal class size. Interestingly, perplexity significantly deteriorates when the number of classes exceeds the optimal number decided by MefMax.


\begin{table}[t]
\small
\begin{tabular}{lccccc}
\toprule
Model        & \begin{tabular}[c]{@{}c@{}}\footnotesize{Decoupled}\\ \footnotesize{Decoding} \end{tabular}  &KLD   &   MSJ   & SB &  Uniq \\ \hline
\rule{0pt}{2.4ex} \multirow{2}{0cm}{\textbf{F$^{2}$-Softmax}}\hspace{3em}  & O &0.13  & 52.4  & 76.1  & 25.2k \\
 & $\times$&0.31  & 47.4    & 81.9      & 22.6k \\
 \hline
\rule{0pt}{2.4ex} \textbf{MLE}     & $\times$   & 0.34  & 47.3    &  83.1     & 22.4k     \\
\bottomrule
\end{tabular}
\caption{Ablation study on the decoupled decoding. MSJ and SB stand for MS-Jaccard and Self-BLEU, respectively. Scores are measured on bigram level.}
\label{wodd}
\end{table}

\subsubsection{Ablation on Decoupled Decoding} Decoupled decoding formalized in Eq. (\ref{dd}) fully leverages the benefits of F$^2$-Softmax by sequentially performing the frequency class prediction and token generation tasks. Table \ref{wodd} reports the results from an ablation test on the decoupled decoding. To ablate the decoupled decoding, we use a full posterior in Eq. (\ref{fs}). We observe significant performance gaps, meaning that the decoupled decoding is an indispensable component of our model. Notably, even without the decoupled decoding, our model maintains better performances than the MLE baseline.

\begin{table*}[t]
\small
	\begin{tabularx}{\textwidth}{l X c}
\toprule[0.3ex]
		\textbf{Model} & \textbf{Texts} & \textbf{VR} (\%) \\

\toprule[0.15ex]
		Prefix  & = Robert Boulter = Robert Boulter is an English film, television and theatre actor. He had a guest-starring role on the television series The Bill in 2000. This was followed by a starring role in the play Herons written & \\ \hline\hline
		\textbf{MLE}  & for the television series The Bill in 2001. \textbf{\textcolor{green}{Boul}}ter was cast in the role of Robert \textbf{\textcolor{green}{Boul}}ter on the television series The Bill in 2002. The character was first cast in the role of Robert \textbf{\textcolor{green}{Boul}}ter in the 2003 film The Bill in a series of short stories about the life of a man.   = = Personal life = =   \textbf{\textcolor{green}{Boul}}ter is a \textbf{\textcolor{cadmiumred}{vegetarian}} & 6.7 \\ \hline
		\textbf{FACE}  & by John \textbf{\textcolor{cadmiumyellow}{Landis}} in 2001 . His role in the play was also played by John Lloyd , who played a fictional character in The Bill . \textbf{\textcolor{green}{Boul}}ter played a leading role in The Bill in 2002 and played a leading role in The Bill in 2004 . In 2006 he starred as a supporting actor in The Bill , a film adaptation of John Wayne 's play The \textbf{\textcolor{fuchsiapink}{Nut}}ty Professor in which & 4 \\ \hline
		\multirow{2}{0cm}{\textbf{UL-token}}  & in 2002 by the BBC. \textbf{\textcolor{green}{Boul}}ter was cast by the BBC, playing a young Englishmen in a \textbf{\textcolor{cadmiumred}{pub}} in the fictional town of Bristol. He appeared in the 2006 BBC drama series, The Bill, in which he was cast as the lead character. He played the role of Peter \textbf{\textcolor{blue}{Gabriel}}, an Englishman who is portrayed in a television serial killer.  \textbf{\textcolor{green}{Boul}}ter is & 5.3 \\ \hline
		\multirow{2}{0cm}{\textbf{UL-token+seq}}  & and directed by David Lloyd.  \textbf{\textcolor{green}{Wins}}let was a director for \textbf{\textcolor{green}{Wins}}let 's comedy-drama, The New \textbf{\textcolor{cadmiumyellow}{Yorker}}. He starred in several of his films, and was also the first actor to appear in multiple films. He played the role in the film 's second part in the 2002 \textbf{\textcolor{blue}{HBO}} series The \textbf{\textcolor{fuchsiapink}{Adventures}} of Bob \textbf{\textcolor{purple}{Marley}} (which premiered in 2001) and was written by David & 8 \\ \hline
		\textbf{F$^{2}$-Softmax}  & by Philip \textbf{\textcolor{cadmiumyellow}{Glass}} in 2003. He co-starred as the male fictionalised father with a woman named Anne \textbf{\textcolor{blue}{Hutchinson}}, an \textbf{\textcolor{cadmiumred}{eccentric}} lawyer who discovers that his family had \textbf{\textcolor{cadmiumred}{vanished}} in the previous film. \textbf{\textcolor{green}{Boul}}ter's performance in A Day in Yellow \textbf{\textcolor{fuchsiapink}{Cell}} is considered a turning point in his career. He received an Academy Award nomination for Best Actor in 2004. \textbf{\textcolor{green}{Boul}}ter appeared in several films & \textbf{9.3} \\
\toprule[0.15ex]
\end{tabularx}
\caption{Generated texts on the Wikitext-103 test set. A prefix from the first batch was selected to avoid cherry-picking. \textbf{VR} denotes the ratio of \textit{very rare} tokens (see Section~\ref{sec:Learning_Balanced_Distribution} for the definition) against the text length. While all colored and bold-faced tokens indicate \textit{very rare} tokens, \textbf{\textcolor{green}{green}} color denotes repeated tokens, and \textbf{\textcolor{cadmiumred}{red}} color is reserved for non-pronoun words.}
		\label{generated}
\end{table*}

\subsection{Qualitative Comparisons}
\label{sec:Qualitative_Comparisons}
To further examine the generation quality, we sample texts from the trained models. Table \ref{generated} compares the generated texts from the same prefix. The results suggest that all trained models are capable of generating texts semantically coherent to the prefix. However, they differ in rare token usage patterns. While our F$^{2}$-Softmax exhibits the highest usage of rare tokens, we observe two issues from the baselines. The first is that models tend to repeat the same rare token across all sentences after its first appearance (MLE). The other issue is that generated rare tokens are mostly pronouns (UL-token-seq). Unlike the baselines, F$^{2}$-Softmax utilizes the broadest range of rare tokens with significantly less, but more likely, repetitions. Further, F$^{2}$-Softmax is shown to be adept at utilizing non-pronoun rare tokens, such as `eccentric' or `vanished'.

\section{Conclusion}
In this paper, we proposed F$^{2}$-Softmax, a simple but effective method for better learning the rich diversity in text. F$^{2}$-Softmax encourages models to diversify text generation by readjusting class formation and motivating models to learn a more balanced token distribution. Quantitative and qualitative analyses validate the diversity-promoting performances of our approach. Since it can be quickly adopted to replace the traditional likelihood objective, we believe in broader applicability of F$^{2}$-Softmax. Thus, future work involves extending the method to other related tasks, such as machine translation and text summarization, and investigating the potential gains from transfer learning.

\section*{Acknowledgments}
This work was supported by Institute of Information \& Communications Technology Planning \& Evaluation (IITP) grant funded by the Korea government (MSIT) (No. 2020-0-01371, Development of brain-inspired AI with human-like intelligence)
\bibliography{anthology,emnlp2020}
\bibliographystyle{acl_natbib}

\clearpage
\appendix
\section{Datasets}
\label{asec:datasets}
\subsection{Melo-Lyrics Data Collection}
Few datasets  have been publicly available for Korean text generation, and none of them has gained public consensus as a benchmark dataset, partly due to their small sample sizes. We collect lyrics data for three rationales. First, we test our model on a language other than English. Second, a large number of songs and lyrics are available. Lastly, lyrics show distributional characteristics at odds with Wikipedia. The crawling session was held between 5\textsuperscript{th} July 2019 to 29\textsuperscript{th} July 2019. After crawling enough data, we discarded those containing more than ten foreign language words, except for English. English was an exception since using English in Korean lyrics is natural and common. We also manually refined the lyrics by deleting noises, including advertisements and unnecessary meta-information about the lyrics writer transcriber. The remaining data consist of roughly 478 thousand lyrics with 51 million words. Indexed data can be downloaded from the url below\footnote{\url{https://drive.google.com/drive/folders/1bXborfoUiaHYU0X_1t-TmnBkClU4ts9O}}. We plan to release the raw data for research purposes only. 
\subsection{Data Statistics}
The number of articles (songs) and containing words for training, test and validation phases are reported in Table \ref{app:datastat}.
\begin{table}[h]
\small
\begin{subtable}{\columnwidth}
\begin{tabular}{l|ccc} \toprule
             & train     & test     & valid     \\\toprule[.15ex]
\# of articles & 28,475 & 60 & 60 \\
\# of words &    113,655,420       & 269,551   &  236,966        
\end{tabular}
\caption{Wikitext-103 dataset}
\end{subtable}

 \vspace{1em}
\begin{subtable}{\columnwidth}
\begin{tabular}{l|ccc} \toprule
             & train     & test     & valid     \\\toprule[.15ex]
\# of songs & 430,837 & 23,935 & 23,935 \\
\# of words &    46,343,239       & 2,566,598   &  2,501,304        
\end{tabular}
\caption{Melo-Lyrics dataset}
\end{subtable}
\caption{Statistics on the datasets}
\label{app:datastat}
\end{table}
\section{Hyperparameter Configurations}
\label{asec:config}
The detailed hyperparameters used are illustrated.
\subsection{Model Hyperparamters}
\label{app:modelconfig}
\begin{table}[h]
\small
\begin{tabular}{l|c|c}\toprule
       \diagbox[innerleftsep=0pt,innerrightsep=0pt, innerwidth=3.3cm]{Hyperparameter}{Dataset}              & \multicolumn{1}{c|}{Wikitext-103} & \multicolumn{1}{c}{Melo-Lyrics} \\ \toprule[0.15ex]
\# of layers            & \multicolumn{2}{c}{12}                                              \\
Hidden dimension     & \multicolumn{2}{c}{512}                                             \\
Projection dimension & \multicolumn{2}{c}{2048}                                            \\
\# of heads          & \multicolumn{2}{c}{8}                                               \\
Head dimmension      & \multicolumn{2}{c}{64}                                              \\
Dropout              & \multicolumn{2}{c}{0.1}                                             \\
Drop attention       & \multicolumn{2}{c}{0}                                               \\\hline
Sequence length      & 1024                              & 512                            \\
Vocabulary size      & 30,000                              & 40,000                            \\ 
Total \# of parameters      & 69.0M                              & 76.5M
                            
\end{tabular}
\caption{Model hyperparameter settings}
\label{app:modelconfig_table}
\end{table}
Table \ref{app:modelconfig_table} reports the detailed list of model hyperparameters. The dropout and drop attention ratios are chosen from a set \{0, 0.1\} based on validation losses. Sequence length is selected from a set \{512, 1024\}. We assigned 10,000 more vocabularies to training models on the Melo-Lyrics dataset, illuminating the characteristics of Korean language where words with varying forms may have similar meanings. 
\subsection{Training Hyperparamters}
\label{app:trainingconfig}
\begin{table}[h]
\small
\begin{tabular}{l|c|c}\toprule
       \diagbox[innerleftsep=0pt,innerrightsep=0pt, innerwidth=3.3cm]{Hyperparameter}{Dataset}              & Wikitext-103 & Melo-Lyrics \\ \toprule[0.15ex]
Batch size            & 8 & 16                                             \\
Learning rate     & 0.0001 & 0.0002                                             \\
Finetuning LR & 0.00001 & 0.00002                                         \\
Finetuning step & 1500 & 1500                                            \\
Gradient clipping & 0.25 & 0.25                                           \\
Weight decay          & 0.001 & 0                                            \\
Optimizer          & Adam & Adam                                            \\
\hspace{1em} - $\beta_1$          & 0.9 & 0.9                                            \\
\hspace{1em} - $\beta_2$          & 0.999 & 0.999                                            \\
\hspace{1em} - $\epsilon$          & 1e-8 & 1e-8                                            \\
\end{tabular}
\caption{Training hyperparameter settings}
\label{app:trainingconfig_tab}
\end{table}

Table \ref{app:trainingconfig_tab} shows the training configurations. The learning rate, gradient clipping norm, weight decay are selected from a set, \{0.00005, 0.0001, 0.00015, 0.0002, 0.00025\}, \{0.25, 5.0\}, \{0, 0.001, 0.0001\}, respectively. Batch sizes are chosen to accommodate the GPU memory constraint. We use default Adam configurations in PyTorch. Finetuning learning rate, selected from a set \{0.00001, 0.00002\}, is used to finetune UL-token-seq and FACE. Of the four variants of FACE, we use FACE-OPR, which reportedly performs best.

\section{Transfer Learning}
\label{app:tl}
Pre-trained language models are widely used for downstream applications by adapting them to domain-specific distributions. Significant gains can be realized if F-$^2$ softmax is successfully applied to fine-tune pre-trained language models, as most pre-trained language models are trained with standard softmax function. To validate our approach on transfer learning settings, we pre-train a language model on news datasets consisting of 10GB of articles and fine-tune the model on the Melo-Lyrics dataset. The results are shown in Table \ref{tr}. The transferred model both increased the quality and diversity of the generation. However, the transferred model exhibits narrower vocabulary usage. We conjecture it is arisen by the vocabulary mismatch between datasets used for pre-training and fine-tuning. We plan to further investigate the vocabulary mismatch problem.

\begin{table}[t]
\small
\centering
\begin{tabular}{lcccc}
\toprule
Model         & PPL   &   MSJ   & SB &  Uniq \\ \hline \vspace{0.3em}
From scratch   & 13.2  & 52.4    &  76.1     & 25.2k \\
 \hline
Transferred     & 9.8  & 56.1    &  74.7     & 23.9k     \\
\bottomrule
\end{tabular}
\caption{Evaluation results on transfer learning.}
\label{tr}
\end{table}

\begin{table*}[h]
\centering
\small
\begin{subtable}{\textwidth}
\centering
\begin{tabular}{lllll}
\toprule
\textbf{Winner} && \textbf{Loser} &\hspace{1ex} \textbf{W-L} & \textbf{Dataset} \\
\toprule[0.15ex]
\textbf{F$^{2}$-Softmax} & \multirow{10}{*}{\textit{beats}} \hspace{1ex}  & \textbf{MLE} &\hspace{1ex} 6-1 & \multirow{10}{*}{\textbf{Wikitext-103}}\\
\textbf{F$^{2}$-Softmax} &  & \textbf{FACE} &\hspace{1ex} 6-1 &  \\
\textbf{F$^{2}$-Softmax} &  & \textbf{UL-token} &\hspace{1ex} 6-1 &\\
\textbf{F$^{2}$-Softmax} &  & \textbf{UL-token-seq} &\hspace{1ex} 5-2 & \\
\textbf{FACE} &  & \textbf{MLE} &\hspace{1ex} 4-3 &  \\
\textbf{UL-token} &  & \textbf{FACE} &\hspace{1ex} 4-3 & \\
\textbf{UL-token-seq} &  & \textbf{FACE} &\hspace{1ex} 4-3 & \\
\textbf{UL-token} &  & \textbf{MLE} &\hspace{1ex} 5-2 & \\
\textbf{UL-token-seq} &  & \textbf{MLE} &\hspace{1ex} 4-3 &  \\
\textbf{UL-token} &  & \textbf{UL-token-seq} &\hspace{1ex} 4-3 & \\ \hline
\textbf{F$^{2}$-Softmax} &  \multirow{10}{*}{\textit{beats}}  & \textbf{UL-token+seq} &\hspace{1ex} 4-3 & \multirow{10}{*}{\textbf{Melo-Lyrics}} \\
\textbf{F$^{2}$-Softmax} &  & \textbf{UL-token} &\hspace{1ex} 4-3 &  \\
\textbf{F$^{2}$-Softmax} & & \textbf{MLE} &\hspace{1ex} 4-3 &  \\
\textbf{FACE} &  & \textbf{F$^{2}$-Softmax} &\hspace{1ex} 4-3 &  \\
\textbf{FACE} &  & \textbf{MLE}  &\hspace{1ex} 4-3 &  \\
\textbf{UL-token} &  & \textbf{FACE}  &\hspace{1ex} 4-3 &  \\
\textbf{FACE} &  & \textbf{UL-token+seq}  &\hspace{1ex} 4-3 &  \\
\textbf{MLE} &  & \textbf{UL-token} &\hspace{1ex} 4-3 &  \\
\textbf{MLE} &  & \textbf{UL-token+seq} &\hspace{1ex} 4-3 &  \\
\textbf{UL-token} &  & \textbf{UL-token+seq} &\hspace{1ex} 4-3 &  \\


\end{tabular}
\subcaption{Results on pair-wise evaluations between models.}
\end{subtable}

 \vspace{2em}
\begin{subtable}{\textwidth}
\centering
\begin{tabular}{llll}
\toprule
\textbf{Rank} & \textbf{Model} &\hspace{1ex} \textbf{Wins} & \textbf{Losses} \\
\toprule[0.15ex]
1&\textbf{F$^{2}$-Softmax} & 7& 1 \\
2&\textbf{UL-token} & 5& 3 \\
3&\textbf{FACE} & 4& 4 \\
\multirow{2}{*}{4}&\textbf{MLE} & 2& 6 \\
&\textbf{UL-token-seq} & 2& 6 \\

\end{tabular}
\subcaption{Model ranking based on the number of wins in pair-wise evaluations.}
\end{subtable}
\caption{Analysis on the greedy sampling results.}
\label{app:comp}
\end{table*}
\begin{table*}[!t]
\small
\centering
\label{app:deterministi}
\begin{subtable}{\textwidth}
\centering
\begin{tabular}{l|l|l|lll|lll|lll|l|l}
\toprule[0.3ex]
                             \multirow{2}{0cm}{\diagbox[innerwidth=1.8cm]{Models}{Metrics}}        & \textbf{PPL} &  \textbf{KLD} & \multicolumn{3}{c|}{\textbf{MS-Jaccard}} & \multicolumn{3}{c|}{\textbf{Self-BLEU}} & \multicolumn{3}{c|}{\textbf{Distinct}} & \textbf{Rep} & \textbf{Uniq} \\
                             
                             & & & \multicolumn{1}{c}{$n$=1}  & \multicolumn{1}{c}{$n$=2} &\multicolumn{1}{c|}{$n$=3}  & \multicolumn{1}{c}{$n$=1}  & \multicolumn{1}{c}{$n$=2} &\multicolumn{1}{c|}{$n$=3} &\multicolumn{1}{c}{$n$=1}  & \multicolumn{1}{c}{$n$=2} &\multicolumn{1}{c|}{$n$=3}&& \\
                                     \toprule

\multicolumn{14}{l}{top-\textit{k}=3} \\
\toprule
\textbf{MLE}      & \bf{24.7}  & 1.51 & 52.1& 35.6 & 24.3 & 93.4  &  83.2 &  69.7  & 45.1  & 71.9  & 83.0 & 0.67 & 8.48k \\
\textbf{FACE}     & 29.7  & 1.26 &  53.3& 33.2 & 21.7  & 92.3 & 76.5  & 57.6   & 53.4  & 77.1 & 85.5 & 2.1 & 10.3k         \\
\textbf{UL-token} & 25.8  & 1.28 &   54.3& 36.7 & 24.8 &  93.7 &     82.4   &  68.1 & 50.0   &  77.3 & 87.1 & 0.39 & 10.2k       \\
\textbf{UL-token+seq}  & 27.5 & 1.33  & 50.6 & 35.4 &  23.5  & \bf{95.3}    &  83.4  &  66.9   & 57.6    &  86.6 & 94.2  & \bf{0.11}  &  10.6k  \\
\hline
\textbf{F$^{2}$-Softmax}   & 25.6 & \bf{0.62} &   \bf{67.4}&  \bf{42.4}  & \bf{26.4} &  93.3   &  \bf{71.9}  & \bf{48.1} &  \bf{65.7}   &  \bf{89.7}  & \bf{94.4}      & 0.33 & \bf{15.7k}    \\

\toprule
\noalign{\vskip 2mm}    
\multicolumn{14}{l}{top-\textit{k}=5} \\
\toprule
\textbf{MLE}    & \bf{24.7} & 1.34 & 55.9&	37.9&	25.6
&	94.1	&82.8	&67.7
&	50.0&	78.1&	87.8
& 0.73 & 9.2k  \\
\textbf{FACE}  & 29.7 & 1.08& 53.2& 34.1&	22.0&	93.2
   & \bf{77.0}&	56.2&	58.3
   & 82.4&	89.4&	0.9 & 11.6k       \\
\textbf{UL-token}    & 25.8 &  1.11 &    57.6&	38.7&	25.8
 &93.9&	81.1&	64.4& 54.6
	&82.0	&90.3
&	0.55 & 12.1k       \\
\textbf{UL-token+seq} & 27.5 & 1.07& 55.8& 36.5&	23.7&	\bf{95.8}
   & 82.1&	62.5&	61.4
   & 90.4&	\bf{96.0}&	\bf{0.09} & 12.2k       \\
\hline
\textbf{F$^{2}$-Softmax}     & 25.6 &   \bf{0.59} & \bf{67.3}	&\bf{41.9}&	\bf{25.9}
& 93.4	&70.6&	\bf{45.9}
&\bf{67.8}	&\bf{90.9}	&95.0
& 0.22 & \bf{16.2k}    \\
\toprule
\noalign{\vskip 2mm}    
\multicolumn{14}{l}{top-\textit{k}=10} \\
\toprule
\textbf{MLE}    & \bf{24.7} & 1.12 & 60.2 & 40.6&	\bf{27.0}&	94.0 & 80.9&	63.1&	54.7
 & 83.0&	91.0 & 0.89 & 10.3k  \\
\textbf{FACE}  & 29.7 & 1.02 & 53.6& 33.9&	21.4&	93.1
   & \bf{74.4}&	\bf{50.6}&	63.5
   & 87.2&	92.6&	0.61 & 12.0k       \\
\textbf{UL-token}    & 25.8 &  0.95 &    62.0 & 41.1&	26.9&	94.2
 & 79.0&	59.4&	59.3
 &  86.5&	93.1&	0.61 & 12.3k       \\
\textbf{UL-token+seq} & 27.5 & 0.93   & 60.9 & 39.3& 25.0& \bf{95.2}	 &  78.0&	55.0&	65.9
 &  92.9&	96.8& \bf{0.05} & 12.9k      \\
\hline
\textbf{F$^{2}$-Softmax}     & 25.6 &   \bf{0.57} & \bf{67.4} &  \bf{41.8}&	25.6& 93.5
   & 69.5 &	44.2&	\bf{69.1}
  & \bf{91.5}&	\bf{95.2}& 0.22 & \bf{16.7k}    \\

\toprule
\noalign{\vskip 2mm}
\multicolumn{14}{l}{top-\textit{k}=20} \\
\toprule
\textbf{MLE}    & \bf{24.7} & 0.95 & 63.5&	\bf{42.4}&	\bf{27.8}&	94.2&	78.9&	58.8&	58.9&	86.5& 93.0 & 0.39 & 11.5k  \\
\textbf{FACE}  & 29.7 & 0.96 & 53.2&	33.3&	20.6
&	93.5&	72.6&	46.4&
67.6&	90.2&	94.2&	0.33 & 12.0k       \\
\textbf{UL-token}    & 25.8 &  0.79 &    65.5&	42.4&	27.3&	94.2&	76.5&	54.5&	63.3&	89.4&	94.6&	0.55 & 13.6k       \\
\textbf{UL-token+seq} & 27.5 & 0.80   & 64.8&	41.0&	25.4&
\bf{95.2}&	\bf{74.9}&	\bf{49.4}&
69.9&	94.4&	97.2
& 0 & \bf{14.1k}      \\
\hline
\textbf{F$^{2}$-Softmax}     & 25.6 &   \bf{0.54} & \bf{67.2}&	41.2&	25.2
& 93.4&	68.2&	42.3
&	\bf{69.8}&	\bf{91.9}	&\bf{95.3}
& \bf{0.16} & 17.1k    \\

\toprule
\toprule

Human                         & - & -  & - &    - &   -   &   95.2 &  74.1   & 50.9  & 69.1& 92.1  & 95.8       & 0.11 & 15.2k  \\

\end{tabular}
\subcaption[]{Wikitext-103}
\end{subtable}
\begin{subtable}{\textwidth}
 \hspace{2em}
\centering
\begin{tabular}{l|l|l|lll|lll|lll|l|l}
\toprule[0.3ex]
                             \multirow{2}{0cm}{\diagbox[innerwidth=1.8cm]{Models}{Metrics}}        & \textbf{PPL} &  \textbf{KLD} & \multicolumn{3}{c|}{\textbf{MS-Jaccard}} & \multicolumn{3}{c|}{\textbf{Self-BLEU}} & \multicolumn{3}{c|}{\textbf{Distinct}} & \textbf{Rep} & \textbf{Uniq} \\

                             & & & \multicolumn{1}{c}{$n$=1}  & \multicolumn{1}{c}{$n$=2} &\multicolumn{1}{c|}{$n$=3}  & \multicolumn{1}{c}{$n$=1}  & \multicolumn{1}{c}{$n$=2} &\multicolumn{1}{c|}{$n$=3} &\multicolumn{1}{c}{$n$=1}  & \multicolumn{1}{c}{$n$=2} &\multicolumn{1}{c|}{$n$=3}&& \\
                             \toprule
                                     \toprule

\multicolumn{14}{l}{top-\textit{k}=3} \\
\toprule
\textbf{MLE}    & \bf{13.1} & 0.39 & 64.6&  44.9 & 31.8 & \bf{94.7} & 77.7 &	59.5	& 31.8
 &43.3&	51.3 &	11.7 & 22.0k  \\
\textbf{FACE}  & 13.9 & 0.45 & 59.4 &40.6 &	28.9 &	94.9
   & 77.5 &	60.1&	34.5
   &47.5 &	56.7  & 8.0 & 22.5k       \\
\textbf{UL-token}    & 13.8 &  0.57 &    56.2& 40.1 & 28.7 & 96.1 &82.3 &	67.3 & 35.5
 &  48.2	& 55.9	& 6.2 & 21.2k       \\
\textbf{UL-token+seq} & 16.6 & 0.66   & 50.7& 36.1	& 25.4& 	97.5
 &  86.1 &	69.9 &	41.4 & \bf{68.1} &	\bf{82.4}&	
 \bf{1.1} & 20.6k      \\
\hline
\textbf{F$^{2}$-Softmax}     & 13.2 &   \bf{0.20} & \bf{75.1}& \bf{51.1} & \bf{35.0}   &   95.7  & \bf{75.4} & \bf{52.5} &	\bf{44.5} &
   57.2 &	63.8 & 11.3 & \bf{24.0}k    \\
   
   \toprule 
\noalign{\vskip 2mm}    
    \multicolumn{14}{l}{top-\textit{k}=5} \\
\toprule
\textbf{MLE}    & \bf{13.1} & 0.39 & 64.2&	45.1	&31.8&
95.7&	80.0&	61.9
&37.2&	51.8	&61.3

  & 0.3 & 22.8k  \\
\textbf{FACE}  & 13.9 & 0.44& 59.8& 40.7&	28.8&	95.8
   & 79.5&	61.7&	39.8
   & 55.4&	65.7&	4.5 & 22.6k       \\
\textbf{UL-token}    & 13.8 &  0.53 &  57.8	&41.3&	29.2&
97.0&	83.8&	68.2
&42.1	&57.2&	65.9
& \bf{2.6} & 21.6k       \\
\textbf{UL-token+seq} & 16.6 & 0.59& 53.3& 37.4&	25.8&	\bf{97.9}
   & 86.1&	68.0&	48.6
   & \bf{76.8}&	\bf{88.5}&	0.5 &20.7k       \\
\hline
\textbf{F$^{2}$-Softmax}     & 13.2 &   \bf{0.19} &  \bf{76.1}	&\bf{51.7}	&\bf{35.1}&
96.3	&\bf{76.2}&	\bf{52.7}&

\bf{50.1}&	63.9&	70.4
 & 7.5 & \bf{24.3k}    \\

\toprule
\noalign{\vskip 2mm}    
\multicolumn{14}{l}{top-\textit{k}=10} \\
\toprule
\textbf{MLE}    & \bf{13.1} & 0.37 &  65.3& 46.0 & 32.1 & 96.7 & 82.1 & 63.3 & 46.5 & 64.3 & 74.2   & 2.4 & 22.1k  \\
\textbf{FACE} & 13.9 & 0.42  &59.5 & 41.1     & 28.6  &  96.7   & 81.3     & 62.0 & 48.9 & 67.4 & 77.5  & \bf{2.1} & 22.6k      \\
\textbf{UL-token}    & 13.8 &  0.46 &    60.3&42.8 & 29.7 & \bf{97.7} & 84.6 & 67.2 & 52.4 &  69.6 & 77.7 & 1.1 & 22.2k       \\
\textbf{UL-token+seq}  & 16.6 & 0.51 & 55.6 & 38.5 & 25.9     & 98.4   & 85.2 & 63.4   & \bf{58.4}   & 85.6 & 93.3 &  0.1 & 21.3k       \\
\hline
\textbf{F$^{2}$-Softmax}     & 13.2 &   \bf{0.16} &  \bf{77.5} & \bf{52.2} & \bf{34.9}   &  97.0 & \bf{76.5}   & \bf{51.9}  & 57.4 & \bf{72.1}  & \bf{78.0} & 4.3 & \bf{24.7k}    \\

\toprule
\noalign{\vskip 2mm}    
\multicolumn{14}{l}{top-\textit{k}=20} \\
                                     
\toprule
\textbf{MLE}      & \bf{13.1}  & 0.34 & 67.6&47.3 & 32.4  & \bf{97.5}  &  83.1 &  62.2 &  56.3 & 75.5  & 84.0 &  1.1 & 22.4k \\
\textbf{FACE}     & 13.9  & 0.39 & 60.3 & 41.5 & 28.3  & \bf{97.5}  & 82.0    & 60.2  & 58.6  & \bf{78.1}  & 86.5  &  0.9      & 22.7k   \\
\textbf{UL-token} & 13.8  & 0.39 &62.4 &  43.6 & 29.6 & 98.2 &84.0  & 64.2 &  61.0   &  78.3 & 84.7  & 0.3 & 22.8k       \\
\textbf{UL-token+seq}  & 16.6 & 0.45   &58.3& 39.4 & 25.7  & 98.7 & 79.1   & 57.1   & 67.2   &  90.7  & 95.0  &   0.07 & 22.3k      \\
\hline
\textbf{F$^{2}$-Softmax}   & 13.2 & \bf{0.13} &    \bf{78.8}& \bf{52.4} & \bf{34.4}  &\bf{97.5}   &  \bf{76.1} &   \bf{50.3} &  \bf{64.0}   &  78.7 &  \bf{83.9} & \bf{2.5} & \bf{25.2k}    \\

\toprule
\toprule
Human                         & - & -  & - &    - &   -   &    97.5& 76.6  & 53.2 &  62.5 &  77.5   & 82.5      & 2.1 & 27.9k  \\

\end{tabular}
\subcaption[]{Melo-Lyrics}
\end{subtable}
\caption{Evaluation results on different top-\textit{k}.}
\label{app:topk}
\end{table*}

\begin{table*}[t]
\small
	\begin{tabularx}{\textwidth}{l X c}
\toprule[0.3ex]
		\textbf{Model} & \textbf{Texts} & \textbf{VR} (\%) \\

\toprule[0.15ex]
		Prefix  & = Clayton Kershaw =   Clayton Edward Kershaw ( born March 19 , 1988 ) is an American professional baseball pitcher for the Los Angeles Dodgers of Major League Baseball ( MLB ). A left @-@ handed starting pitcher, Kershaw & \\ \hline\hline

		\textbf{MLE}  & began his career at \textbf{\textcolor{cadmiumred}{shortstop}} for the Los Angeles \textbf{\textcolor{green}{Dodgers}} in 2006. He spent his entire career with the \textbf{\textcolor{green}{Dodgers}} before moving to the minor leagues in 2006 , where he spent two seasons and played for the New York Yankees before \textbf{\textcolor{cadmiumred}{retiring}} in 2008. He was traded to the San Francisco Giants in 2011 , where he spent two seasons with the St. Louis Cardinals.  Ker\textbf{\textcolor{green}{shaw}} was a two @-@ time MLB player in 2006. & 6.0 \\ \hline
		
\textbf{F$^{2}$-Softmax}  & set a team record with eight National League runs scored in 110 innings pitched in his first three seasons. A right @-@ handed pitch attempts to make the \textbf{\textcolor{cadmiumred}{pitching}} \textbf{\textcolor{cadmiumred}{swing}} start , but his starting place as an extra @-@ major leagues team goes so far as to \textbf{\textcolor{cadmiumred}{boost}} his chances of victory. He is best known for his efforts in both \textbf{\textcolor{green}{pitching}} and baseruster fastball ( known as the " Wild Wings " ) and his willingness to throw \textbf{\textcolor{cadmiumred}{pitches}} into \textbf{\textcolor{cadmiumred}{bats}}. & 8.9 \\
\toprule[0.15ex]
\vspace{0.5cm}
\end{tabularx}

	\begin{tabularx}{\textwidth}{l X c}
\toprule[0.3ex]
		\textbf{Model} & \textbf{Texts} & \textbf{VR} (\%) \\

\toprule[0.15ex]
		Prefix  &  = Hed PE =   Hed PE, also known as ( hed ) Planet Earth and stylized as (  ), is an American rock band from Huntington Beach, California. Formed in 1994, the band performs a style of & \\ \hline\hline

		\textbf{MLE}  & music called the  and has released two studio albums , The  of which were released as the group's second studio album in 1995. The band has released four studio albums, two albums, two albums and two albums, as well as a live album of their own.   = = History = =   \textbf{\textcolor{green}{Hed}} \textbf{\textcolor{green}{Pe}} was born in \textbf{\textcolor{green}{Huntington}} Beach, California, to a family of American immigrants, in the town of \textbf{\textcolor{green}{Huntington}} Beach. & 4.4 \\ \hline
		
\textbf{F$^{2}$-Softmax}  & jazz opera , jazz and \textbf{\textcolor{cadmiumred}{avant}} @-@ \textbf{\textcolor{cadmiumred}{garde}} jazz for the 1990s. It started under \textbf{\textcolor{fuchsiapink}{Epic}} Records in 2003 and received a \$ 100 @,@ 000 grant from Columbia Records. The New York Knicks released their second album The Way \textbf{\textcolor{cadmiumyellow}{Beyond}} the \textbf{\textcolor{blue}{Sunset}}, in March 2003. \textbf{\textcolor{cadmiumred}{Led}} by vocalist Eddie Murphy, \textbf{\textcolor{purple}{Hedwig}} grew up in Portland and worked as a \textbf{\textcolor{cadmiumred}{consulting}} keyboardist for Mercury Records, then joined Mercury Records in 1999. & 12.3 \\
\toprule[0.15ex]
\vspace{0.5cm}
\end{tabularx}

	\begin{tabularx}{\textwidth}{l X c}
\toprule[0.3ex]
		\textbf{Model} & \textbf{Texts} & \textbf{VR} (\%) \\

\toprule[0.15ex]
		Prefix  &  = Little Gidding ( poem ) =   Little Gidding is the fourth and final poem of T. S. Eliot's Four Quartets, a series of poems that discuss time, perspective, humanity, and salvation. It was first published in & \\ \hline\hline

		\textbf{MLE}  & the spring of \textbf{\textcolor{blue}{1838}}.  The poem is the first of \textbf{\textcolor{green}{Eliot}}'s poems to be published by the public. The first of these is `` The Great ", the first poem to be translated into the English language. \textbf{\textcolor{green}{Eliot}} wrote in his \textbf{\textcolor{cadmiumred}{diary}} of the poem, `` The Great ", that the poem ``was written in the morning and afternoon of the night ". & 4.6 \\ \hline
		
\textbf{F$^{2}$-Softmax}  & 
August 1969 with poetry about progress in \textbf{\textcolor{cadmiumred}{salvation}}, and was praised by critics for its lyrical content. The poem was positively received by those who criticized its emphasis on \textbf{\textcolor{cadmiumred}{contemp}}ation, lyrical content, and political themes, while \textbf{\textcolor{green}{Eliot}}'s critical \textbf{\textcolor{cadmiumred}{essays}} about the poetry and short story were criticized.  The poem is set in light of the faith, which focuses on the desire for freedom, change in nature, and self @-@ \textbf{\textcolor{cadmiumred}{esteem}}. & 6.0 \\
\toprule[0.15ex]
\vspace{0.5cm}
\end{tabularx}

	\begin{tabularx}{\textwidth}{l X c}
\toprule[0.3ex]
		\textbf{Model} & \textbf{Texts} & \textbf{VR} (\%) \\

\toprule[0.15ex]
		Prefix  & = Olmec colossal heads =   The Olmec colossal heads are at least seventeen monumental stone representations of human heads sculpted from large basalt boulders. The heads date from at least before 900 BC and are a & \\ \hline\hline

		\textbf{MLE}  & type of stone carved from the walls of the Olmec colossal head of a small stone temple. It has a wide and narrow head and has a long and wide head that is a wide, broad face. The heads are long and long and \textbf{\textcolor{cadmiumred}{triangular}} with rounded \textbf{\textcolor{cadmiumred}{corners}}. The head is long and broad with long, curved \textbf{\textcolor{green}{corners}}. The legs are short and narrow, with short, straight sides, which are not wide. & 3.4 \\ \hline
		
\textbf{F$^{2}$-Softmax}  & broad, highly curved structure that includes the uppermost \textbf{\textcolor{cadmiumred}{skeleton}} of the king @-@ made objects. The body of Olmec colossal is \textbf{\textcolor{cadmiumred}{elongated}} and consists of approximately eight layers on the \textbf{\textcolor{cadmiumred}{obverse}}, followed by two large fragments. The \textbf{\textcolor{cadmiumred}{jaws}} were composed of small bony \textbf{\textcolor{cadmiumred}{projections}} formed by various iron \textbf{\textcolor{green}{skeletons}} and long thin \textbf{\textcolor{cadmiumred}{joints}}. The skull is \textbf{\textcolor{cadmiumred}{unusually}} similar to contemporary humans , with only one \textbf{\textcolor{green}{skeleton}} estimated at 0 @.@ 25 m ( 1 @.@ 5 ft ) across. & 11.1 \\
\toprule[0.15ex]
\end{tabularx}

\caption{Generated samples.}
\end{table*}
\end{document}